\documentclass[conference]{IEEEtran}
\IEEEoverridecommandlockouts

\usepackage{amsmath,amssymb,amsfonts}
\usepackage{algorithmic}
\usepackage{graphicx}
\usepackage{textcomp}
\usepackage{xcolor}
\usepackage[numbers]{natbib}
\usepackage[ruled]{algorithm2e}
\usepackage{subcaption}
\usepackage{wrapfig}
\usepackage{threeparttable}
\usepackage{pifont}
\usepackage{cleveref}
\usepackage{amssymb}
\usepackage{multirow}
\usepackage{booktabs}
\usepackage{pifont}
\usepackage[font=small]{caption}
\def\BibTeX{{\rm B\kern-.05em{\sc i\kern-.025em b}\kern-.08em
    T\kern-.1667em\lower.7ex\hbox{E}\kern-.125emX}}
\begin{document}


\title{Towards Efficient Deep Hashing Retrieval: Condensing Your Data via Feature-Embedding Matching\\
\thanks{$^\dagger$Corresponding author. $^\star$Equal contribution. This work was sponsored in part by the National Natural Science Foundation of China under Grant 62373095, Discipline Innovation Cultivation Program under Grant XKCX202304 and the Fundamental Research Funds for the Central Universities under Grant 2232022G-09.}
}

\author{
    Tao Feng$^{1\star}$\quad 
    Jie Zhang$^{2\star}$\quad  
    Huashan Liu$^{1\dagger}$\quad
    Zhijie Wang$^{1}$\quad
    Shengyuan Pang$^{3}$\\
    $^{1}$ College of Information Science and Technology, Donghua University, Shanghai, China \\
    $^{2}$ Department of Computer Science, ETH Zurich, Zurich, Switzerland \\
    $^{3}$ College of Electrical Engineering, Zhejiang University, Hangzhou, China \\
}

\maketitle

\begin{abstract}
Deep hashing retrieval has gained widespread use in big data retrieval due to its robust feature extraction and efficient hashing process. However, training advanced deep hashing models has become more expensive due to complex optimizations and large datasets. Coreset selection and Dataset Condensation lower overall training costs by reducing the volume of training data without significantly compromising model accuracy for classification task.  In this paper, we explore the effect of mainstream dataset condensation methods for deep hashing retrieval and propose \textbf{IEM} (\textbf{I}nformation-intensive feature-\textbf{E}mbedding \textbf{M}atching), which is centered on distribution matching and incorporates model and data augmentation techniques to further enhance the feature of hashing space. Extensive experiments demonstrate the superior performance and efficiency of our approach.
\end{abstract}


\vspace{-5mm}
\section{Introduction}
In recent years, hashing algorithms have exhibited considerable efficacy in large-scale and high-dimensional data retrieval, owing to their computational efficiency and low storage overhead in practical applications~\cite{cantini2021learning,li2010b,pandey2017general,fan2014cuckoo}. With the evolution of deep learning, there has been a growing interest in combining it with hashing techniques for image retrievals~\cite{cao2017hashnet,fan2020deep,dubey2021decade}, a method known as Deep Hashing Retrieval (DHR).


DHR aims to represent the content of an image using compact hashing code and is achieved by comparing the Hamming distance between the query image and the images of database. The training process of DHR is complex and typically involves multiple optimization objectives~\cite{yuan2020central,jang2021deep,dubey2021decade}. The complexity and time demands of training a state-of-the-art retrieval model have escalated due to multiple optimization objectives and the large-scale nature of real-world datasets. To overcome these challenges, we consider minimizing the amount of training data to ease the demand on computational resources and accelerate the training process. However, a substantial reduction in training data will unavoidably compromise the effect of retrieval model, making it necessary to find a balance between them.

Recent studies have utilized coreset selection~\cite{chen2012super,rebuffi2017icarl,toneva2018empirical} to choose representative samples in order to reduce training costs. However, the effectiveness of coreset selection diminishes when applied to small sample sizes. Besides, Dataset Condensation (DC)~\cite{zhao2020dataset,wang2018dataset,zhao2023dataset,kim2022dataset,cazenavette2022dataset} encapsulates the rich information of the original dataset into a compact synthetic set, enabling a model trained on it to achieve comparable test accuracy to one trained on the full original dataset. DC has been widely applied in classification tasks. In order to reduce the training costs associated with hashing retrieval, it is worth exploring the potential of DC for reducing the retrieval training set. This leads us to the following question:

\textit{\textbf{Question:} Is it possible to adopt existing techniques to develop a dataset condensation method effectively for deep hashing retrieval?}

To address this question, we conduct an empirical study and observe that DC with distribution matching is more effective for deep hashing retrieval tasks (see Section~\ref{empirical}). Based on this, we take perspectives from augmentations on initial network and dataset to further achieve better DC on DHR. We term this approach \textbf{IEM} (\textbf{I}nformation-intensive Feature-\textbf{E}mbedding \textbf{M}atching).

Our contributions can be concluded as follows:
\begin{itemize}
    \item We highlight that this investigation is the first to address dataset condensation in the context of DHR, establishing the initial baselines for this task.
    \item We explore that DC with distribution matching proves to be more effective for DHR compared to other mainstream methods. Based on this, we introduce model and data augmentations, leading to the development of our method IEM, a dataset condensation method for DHR.
    \item Extensive experiments have demonstrated that IEM exhibits exceptional condensation efficiency and can be adapt to existing deep hashing retrieval methods.
\end{itemize}

\section{Dataset Condensation in Deep Hashing Retrieval}\label{empirical}
\vspace{-1mm}
\subsection{Problem Setup}
\vspace{-0.5mm}
The original training set of DHR is $\mathcal{T}=\left\{\left(x_i, y_i\right)\right\}_{i=1}^{|\mathcal{T}|}$, where $x_i \in \mathbb{R}^d$ denotes the real set, and $y_i \in \mathcal{Y}=\{0,1, \ldots, c-1\}$ denotes the corresponding labels of the real data, $c$ represents the number of classes. Our goal is to condense real set into smaller synthetic set $S = \left\{\left(s_j, y_j\right)\right\}_{j=1}^{|\mathcal{S}|}$, where $s_j \in \mathbb{R}^d$ denotes the synthetic training set and $y_i \in \mathcal{Y}$. The number of each class in synthetic data is smaller than the number of each class in real data, therefore $|\mathcal{S}| \ll|\mathcal{T}|$. We define that the deep hashing network as: $f\left(\theta_g, \theta_h\right) \triangleq g\left(\theta_g\right) \circ h\left(\theta_h\right)$, where $h: \mathcal{X} \rightarrow \mathcal{Z}$ is a feature extractor, and $g: \mathcal{Z} \rightarrow \mathcal{V}$ is a hash layer. $\mathcal{X}$, $\mathcal{Z}$ and $\mathcal{V}$ denote the input of image, the output of feature extractor and the hashing code, respectively. The whole network parameters are $\boldsymbol{\theta}=\left\{\theta_g, \theta_h\right\}$.

We denote the optimized weight parameters obtained by minimizing an empirical loss term over the real set $\mathcal{T}$ and synthetic training set $S$ as: 
\begin{align}
    {\theta}^{\mathcal{T}}=\underset{{\theta}^{\mathcal{T}}}{\arg \min } \sum_{\left(x_i, y_i\right) \in \mathcal{T}} l_{hash}\left(f_{{\theta}^{\mathcal{T}}}, x_i, y_i\right), \\
{\theta}^{\mathcal{S}}=\underset{{\theta}^{\mathcal{S}}}{\arg \min } \sum_{\left(s_j, y_j\right) \in \mathcal{S}} l_{hash}\left(f_{{\theta}^{\mathcal{S}}}, s_j, y_j\right),
\end{align} 
where $l_{hash}$ denotes the losses of DHR. We aim to generate the smaller synthetic set $\mathcal{S}$ so that ${\theta}^{\mathcal{T}}$ can obtain comparable retrieval capacity compared with ${\theta}^{\mathcal{S}}$. Following previous works~\cite{jang2021deep,hoe2021one,su2019deep}, we take the mean Average Precision (mAP) as evaluation metrics of retrieval quality. In this case, given the synthetic set $\mathcal{S}$, our objective can be defined as follows:
\begin{equation}\label{eq:distillation_goal}
\mathbb{E}_{\boldsymbol{x} \sim {\mathbf{D}_{qu}}}[\mathbf{m}\left(f\left(x ; {\theta}^{\mathcal{T}}\right)\right)] \simeq \mathbb{E}_{\boldsymbol{x} \sim {\mathbf{D}_{qu}}}[\mathbf{m}\left(f\left(x ; {\theta}^{\mathcal{S}}\right)\right)],
\end{equation}
where the $\mathbf{m}$ denotes the mAP and we employ the same query set $\mathbf{D}_{qu}$ to fairly assess the retrieval quality of both ${\theta}^{\mathcal{T}}$ and ${\theta}^{\mathcal{S}}$. Due to space limitations, we encourage readers to consult the relevant literature about DHR for further information~\cite{dubey2021decade}.

\vspace{-1mm}
\subsection{Distribution Matching is Better to Condense for DHR}\label{dml}
\vspace{-0.5mm}
In order to reduce the training set size of DHR, we explore the effect of coreset selection such as Random (random selection) and Herding~\cite{chen2012super}) and dataset condensation (DSA~\cite{zhao2021dataset} and DM~\cite{zhao2023dataset}), where DSA denotes DC via gradient matching and DM denotes DC via distribution matching. Furthermore, to gain a deeper understanding of how different methods perform in classification versus hashing retrieval, we systematically verify their impacts on these tasks separately. We evaluate the synthetic data of 1/10/50 image(s) per class from CIFAR-10 through ConvNet-3. We employ DHD~\cite{jang2021deep} as the benchmark of DHR.

\begin{figure}[t]
    \centering
    \centering
    \includegraphics[width=0.9\columnwidth]{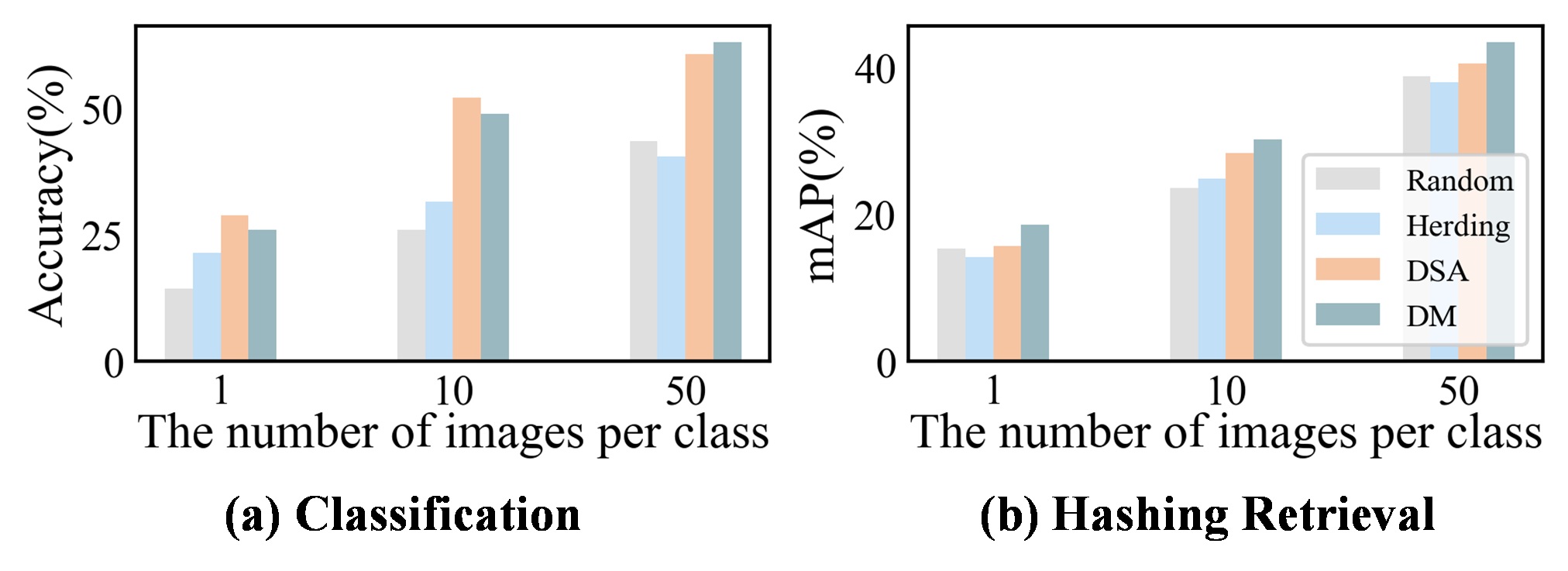}
    \vspace{-2.5mm}
    \captionsetup{font={footnotesize}}
    \caption{\textbf{(a)} The classified performance of condensed data. \textbf{(b)} The retrieval performance of condensed data.}
    \label{emp_1}
    \vspace{-6.5mm}
\end{figure}

As shown in Fig.~\ref{emp_1}, we observe that (1) irrespective of the task— classification or hashing retrieval—coreset selection is markedly less effective than DC when using smaller synthetic data. Notably, Herding even underperforms compared to Random in hashing retrieval. Therefore, to substantially reduce training costs by reducing training data, coreset selection is not a viable approach here. Furthermore, we find that (2) condensation based on distribution matching is more appropriate for retrieval task. Specifically, DSA outperforms DM in classification task, whereas DM leads in retrieval task. Unlike classification, the complex optimization strategies in DHR may lead to increased bias of gradient-matching, though this is an inherent challenge in DHR.

Building on these observations, we intend to apply distribution matching to realize the dataset condensation for the training set of DHR.

\vspace{-1mm}
\section{Method}
\begin{figure}[t] 
    \centering
    \includegraphics[width=0.9\columnwidth]{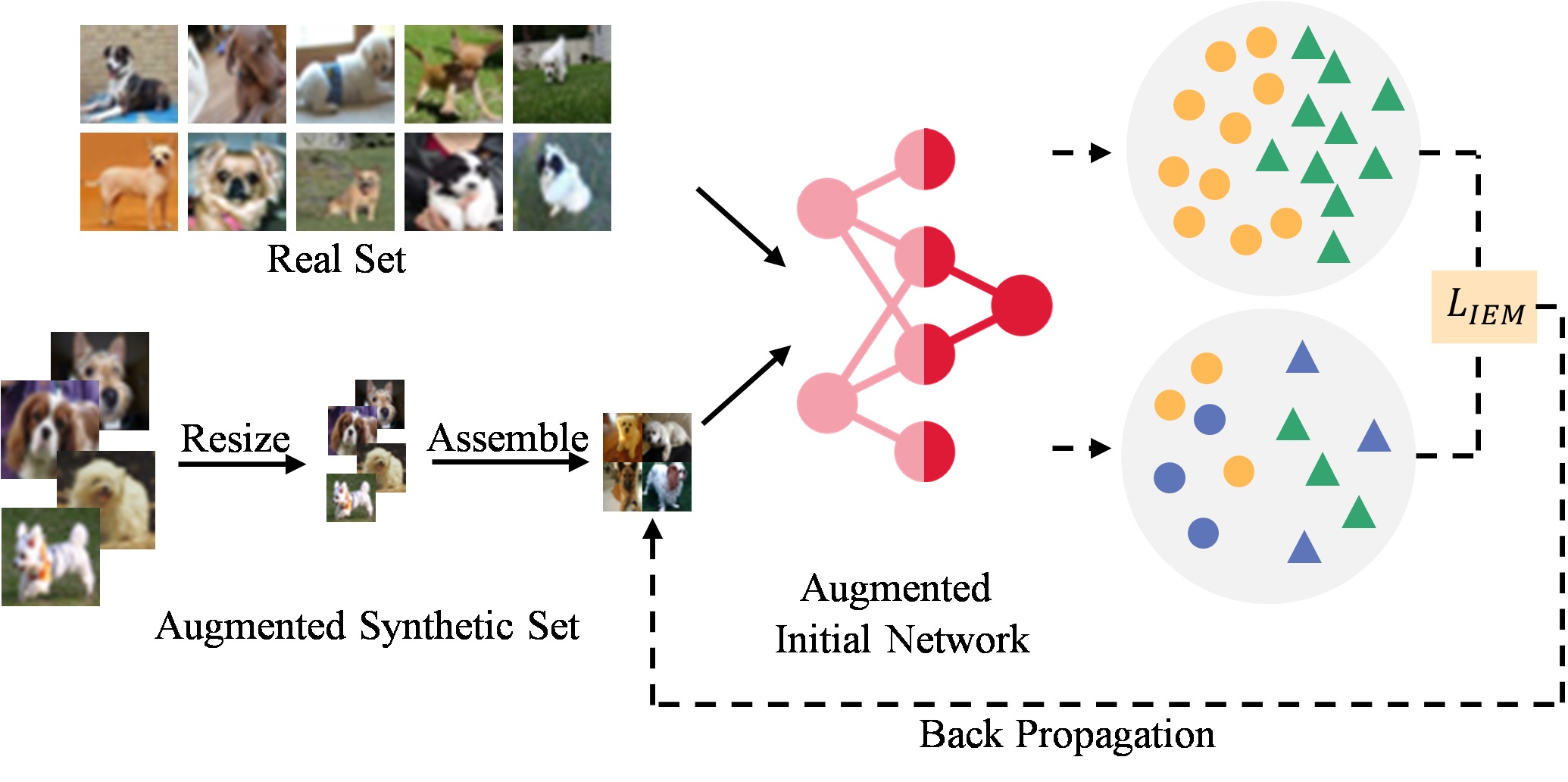}
    \vspace{-1.5mm}
    \captionsetup{font={footnotesize}}
    \caption{The illustration of our proposed method (IEM) for DHR. First, we introduce perturbations to the initialized model and enhance the initialized synthetic data. Subsequently, we updata the synthetic data by feature matching between the distributions of the original and synthetic data.} 
    \label{fig:example-image}
    \vspace{-5.5mm}
\end{figure}

\vspace{-1mm}
In order to develop a dataset condensation method that is compatible with DHR, we utilize distribution matching as in Section~\ref{dml}. However, To fully exploit the benefits of distribution matching in condensing dataset for DHR, we present a method called IEM (\textbf{I}nformation-intensive Feature-\textbf{E}mbedding \textbf{M}atching). The detailed procedure is presented in Fig.~\ref{fig:example-image}.

The core idea of distribution matching is to match the distribution of feature-embedding between real and synthetic set. From the perspective of DHR, high dimensional feature-embedding $\mathcal{Z} = h\left(\mathcal{X},\theta_h\right)$ is projected through hash layer $g\left(\theta_g\right)$ to generate low dimensional continuous hashing code $\mathcal{V}\in \mathbb{R}^K$ (eg:$k<=128$ in almost cases). It is evident that high-quality hashing codes are critical for effective hashing retrieval. Consequently, we aim to transfer the information about the hashing distribution of the original data to the synthetic data from the perspective of DM. However, owing to the reduced semantic information in the continuous hashing code space, we restrict our distribution matching to the feature-embedding space, which contains richer semantic information.

In our scenario, feature-embedding space represents the distribution in the continuous hashing codes space approximatively. Despite this, the feature distribution in a small synthetic dataset is insufficiently representative of the real data distribution, causing a degradation in the effectiveness of distribution matching. Therefore, in order to better condense informative synthetic data for training retrieval model, we must increase the diversity within the feature-embedding space. Our approach is grounded in the following two perspectives:

\textbf{\textit{Network Augmentation.}} Naive DM gets the distribution of feature-embedding through the initial networks and proves the pretrained networks performs not better than initial networks through experiments. However, the state of networks influence the obtained feature-embedding directly~\cite{zhang2022accelerating}. Therefore, we implement a Network augmentation approach to enhance the feature extraction capacity of the initialized network within DM. Inspired by~\cite{nam2022improving,zhang2022accelerating}, we use weight perturbation on networks. We observe that adding perturbations to the pre-trained models is less effective than perturbing the initialized model ${\theta}_{init}$ alone. Hence, we limit the perturbations to the initialized model. The specific process is as follows:
\begin{equation}\label{eq:initial_network}
    \theta_{aug} = \theta_{init} + \alpha \times \mathbf{d}, \quad \mathbf{d} \leftarrow \frac{\mathbf{d} \times \theta_{init}}{\|\mathbf{d}\|_F},
\end{equation}
where $d$ is sampled from a Gaussian distribution $\mathcal{N}(0,1)$, and $\|\cdot\|_F$ represents the Frobenius norm.

\textbf{\textit{Dataset Augmentation.}} In order to substantially increase the characteristic diversity from limited synthetic data, it is essential to consider the synthetic data itself. Spontaneously, we leverage data augmentation to enrich feature embedding. Inspired by IDC~\cite{kim2022dataset}, we utilize mutil-formation to resize the images from the same class and inject these images to each single synthetic data $s_i$ as in Fig.~\ref{fig:example-image}, the detailed process is as follows:
\begin{equation}\label{eq:inject_resize}
s_j^c=\operatorname{Assemble}\left(\operatorname{Resize}\left((s_1^c,\cdots s_i^c),2l/\left | i \right |  \right)\right),
\end{equation}
where the $l$ represents the resolution of real images, $| i |$ represents the number of real data, usually $| i |=4$ for low resolution, and $| i |=6$ for high resolution. As in Eq.~\ref{eq:inject_resize}, we first start with resizing each image of the same class to ($2l/|i| \times 2l/|i|$), and then assemble the resized segments into a complete synthetic image of ($l \times l$). Here we define the augmented synthetic set as:
\begin{equation}\label{eq:new_syn}
S^J = \left\{\left(s_j, y_j\right)\right\}_{j=1}^{|\mathcal{S^J}|}
\end{equation}

After augmentations on both network and dataset, We update the synthetic data by distribution matching loss between real and synthetic set. The loss is as follows:

\begin{equation}\label{eq:loss}
\begin{aligned}
L_{IEM} &= \left\|\frac{1}{\left|T_c^{\mathcal{T}}\right|} \sum_{(\boldsymbol{x}, y) \in T_c^{\mathcal{T}}} \psi_{\theta_{aug}}\left(a_{w}({x}\right)) \right. \\
&\quad - \left. \frac{1}{\left|S_c^{\mathcal{J}}\right|} \sum_{(\boldsymbol{s}, y) \in S_c^{\mathcal{J}}} \psi_{\theta_{aug}}\left(a_{w}({s}\right))\right\|^2
\end{aligned}
\end{equation}

The full details are described in Algorithm~\ref{alg:algorithm1}. During the training process, synthetic data $\mathcal{S}$ is trained over $R$ iterations. Each iteration begins with perturbing the initialized network ${\theta}_{init}$, followed by sampling pairs of real set $T_{c}$ and synthetic data $S_{c}$ batches and augmenting the synthetic data. The mean discrepancy is calculated for each class and aggregated to form the loss $L_{IEM}$. The synthetic data is subsequently updated by minimizing $L_{IEM}$ using stochastic gradient descent with a learning rate $\eta$. The number of iterations $R$ is 200 for CIFAR10 and 300 for ImageNet-subset.

\begin{algorithm}[t]
            \caption{IEM}
            \label{alg:algorithm1}
            \LinesNumbered
            \KwIn{Training set {$\mathcal{T}$}}
            \KwOut{Synthetic set {$\mathcal{S}$}}  
            \textbf{Notation}: Number of classes $N_c$, Perturbation function $\mathcal{P}$, Assemble-Resize function $\mathcal{A_R}$
            Initialize synthetic dataset $S$ \\
            \Repeat{convergence}{
                \For{$r \leftarrow 0$ \textbf{to} $R$}{
                    Perturbation on initial network: $\theta_{aug} = \mathcal{P}(\theta_{init})$
            
                    \For{$c \leftarrow 0$ \textbf{to} $N_{c}$}{
                        Sample an intra-class mini-batch: $T_{c} \sim \mathcal{T}$, $S_{c} \sim S$
            
                        Apply augmentation on $S^J_{c}$: $S^J_c = \mathcal{A_R}(S_{c})$ 
                        Update $S^J_{c} \gets S^J_{c} - \eta \nabla_{c} \left ( L_{IEM} \right )$
                    }
                }
            }
\end{algorithm}

\vspace{-1mm}
\section{Experiments}
\subsection{Implementation Details}
\vspace{-0.5mm}


\textbf{Detailed settings.} In alignment with previous condensation 
methodologies~\cite{kim2022dataset,zhao2023dataset}, we employ ConvNet-3~\cite{sagun2017empirical} for the low-resolution dataset CIFAR10, and ResNetAP-10~\cite{he2016deep} for the high-resolution datasets ImageNet. To assess the performance of condensed datasets in retrieval tasks~\cite{jang2021deep,fan2020deep,hoe2021one}, we train a deep hashing network using condensed data and subsequently test its performance on the query set.  Mean Average Precision (mAP) is employed to measure retrieval performance at hashing code lengths of 32 and 64 bits.

\textbf{Baselines.} To evaluate the effectiveness of IEM in hashing retrieval tasks, we utilize two widely recognized dataset condensation methods as baselines, including (1) gradient matching methods: DSA~\cite{zhao2021dataset} and IDC~\cite{kim2022dataset}, and (2) distribution matching method : DM~\cite{zhao2023dataset}. As for DHR, we utilize the state-of-the-art method DHD~\cite{jang2021deep} as the benchmark.

\begin{table*}[t]
\centering
\captionsetup{font={footnotesize}}
\caption{Comparison of different dataset condensation methods for DHR on CIFAR10 and ImageNet-subset. Img(s)/cls and Ratio (\%) refer to the number of images per class and the size of the synthetic set as a percentage of the total training set, respectively.}
\vspace{-1.5mm}
\label{imagenet-subset}
\begin{tabular}{cccccccccc}
\toprule
                                     & \textbf{Img(s)/cls} & \textbf{Ratio}          & \textbf{Bites} & \textbf{Random} & \textbf{DM} & \textbf{DSA} & \textbf{IDC}   & \textbf{IEM}   & \textbf{Whole set} \\ \midrule
\multirow{6}{*}{\textbf{CIFAR10}}    & \multirow{2}{*}{1}  & \multirow{2}{*}{0.2\%}  & 32             & 14.83           & 18.46       & 15.12        & 23.67          & \textbf{24.21} & 68.15              \\
                                     &                     &                         & 64             & 15.35           & 18.62       & 15.73        & 24.65          & \textbf{25.82} & 70.21              \\
                                     & \multirow{2}{*}{10} & \multirow{2}{*}{2\%}    & 32             & 22.99           & 29.95       & 27.23        & 38.92          & \textbf{45.32} & 68.15              \\
                                     &                     &                         & 64             & 23.54           & 30.23       & 28.32        & 40.32          & \textbf{45.85} & 70.21              \\
                                     & \multirow{2}{*}{50} & \multirow{2}{*}{10\%}   & 32             & 35.47           & 41.12       & 39.83        & 46.78          & \textbf{55.72} & 68.15              \\
                                     &                     &                         & 64             & 38.82           & 43.43       & 40.58        & 46.21          & \textbf{57.49} & 70.21              \\ \midrule
\multirow{4}{*}{\textbf{ImageNet10}} & \multirow{2}{*}{1}  & \multirow{2}{*}{0.77\%} & 32             & 16.46           & 19.84       & 18.63        & 37.21          & \textbf{38.32} & 63.87              \\
                                     &                     &                         & 64             & 17.62           & 20.43       & 18.42        & 39.55          & \textbf{40.32} & 66.05              \\
                                     & \multirow{2}{*}{3}  & \multirow{2}{*}{2.3\%}  & 32             & 20.48           & 21.95       & 20.93        & 45.78          & \textbf{48.69} & 63.87              \\
                                     &                     &                         & 64             & 21.22           & 24.10       & 21.78        & 48.14          & \textbf{52.79} & 66.05              \\ \midrule
\multirow{4}{*}{\textbf{ImageNet20}} & \multirow{2}{*}{1}  & \multirow{2}{*}{0.77\%} & 32             & 10.47           & 10.22       & 10.94        & \textbf{24.89} & 24.23          & 52.05              \\
                                     &                     &                         & 64             & 10.14           & 11.22       & 10.84        & \textbf{26.60} & 25.70          & 57.18              \\
                                     & \multirow{2}{*}{3}  & \multirow{2}{*}{2.3\%}  & 32             & 10.88           & 12.88       & 11.73        & 32.89          & \textbf{33.34} & 52.05              \\
                                     &                     &                         & 64             & 10.71           & 12.96       & 11.78        & 34.48          & \textbf{35.12} & 57.18              \\ \bottomrule
\end{tabular}
\vspace{-6.5mm}
\end{table*}


\vspace{-1mm}
\subsection{Synthetic Set Evaluation.}
\vspace{-0.5mm}

\textbf{CIFAR10.} Following previous works~\cite{zhao2021dataset, zhao2023dataset, kim2022dataset}, we condense CIFAR10 to 1, 10, and 50 image(s) per class. The results provided in Table~\ref{imagenet-subset} indicate that methods such as DSA, DM, and IDC significantly lag behind the performance of the original dataset (Whole set). Nevertheless, IEM exhibits remarkable proficiency across various synthetic data sizes, outperforming IDC by approximately 6\% and 12\% at 10 and 50 images per class, respectively.

\textbf{ImageNet-subset.} We condense ImageNet-subset to 1 and 3 image(s) per class, with each class containing 1200 images. It is noteworthy that we intentionally avoid employing the pre-trained network trained on the entirety of the ImageNet, which ensures a just assessment of the effects for condensation methods. As illustrated in Table~\ref{imagenet-subset},  IEM demonstrates superior performance, outperforming most of the baseline.  However,  IDC achieves comparable results with IEM on ImageNet20. Nonetheless, it is important to acknowledge that the benefits of IDC come at the expense of its own costly condensation process. We provide an in-depth analysis in Section~\ref{Efficiency}.

\vspace{-1mm}
\subsection{Efficiency Comparison.}\label{Efficiency}
\begin{figure}[t] 
    \centering
    \includegraphics[width=0.85\columnwidth]{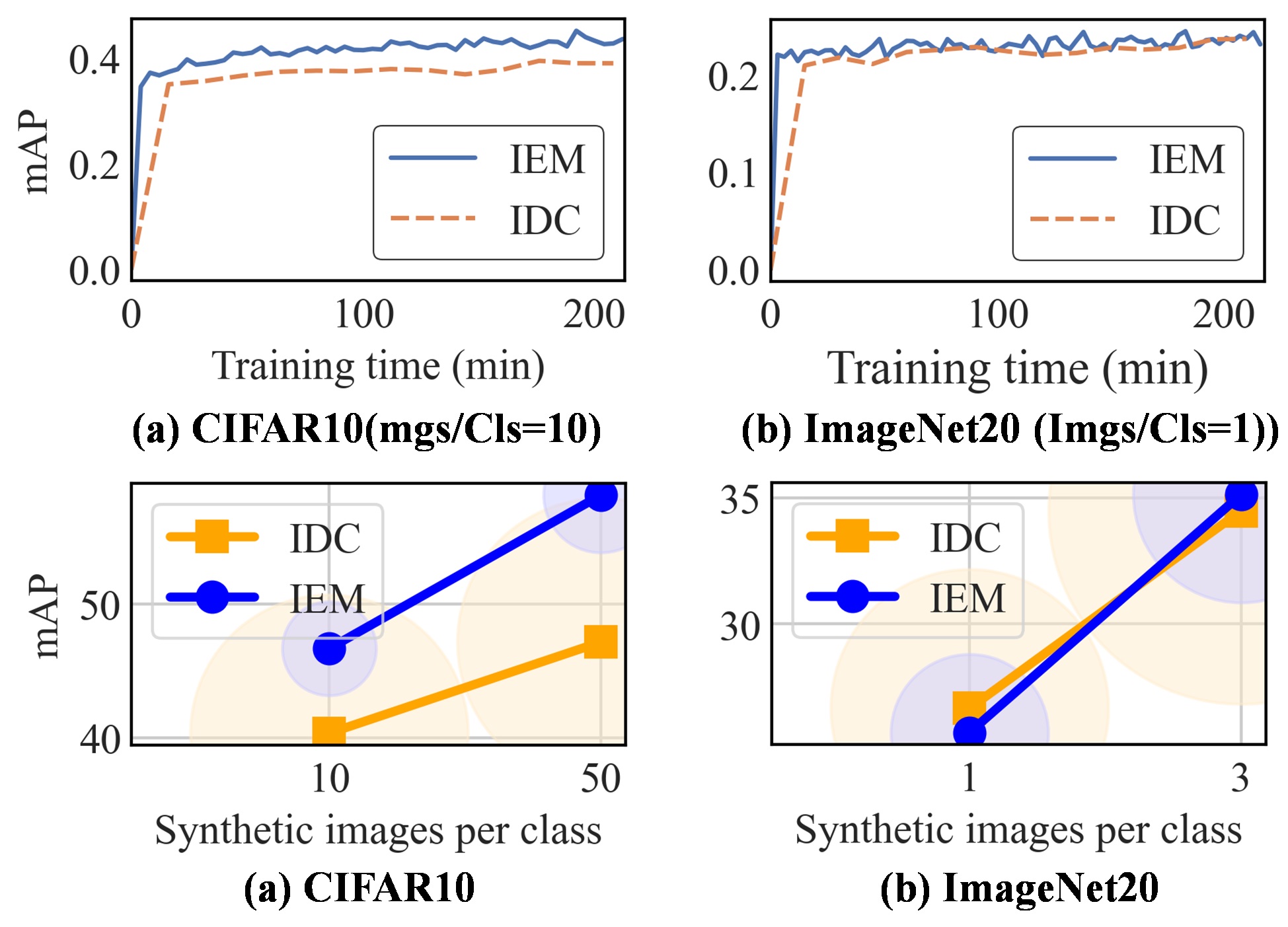}
    \vspace{-1.5mm}
    \captionsetup{font={footnotesize}}
    \caption{Comparison of performance across varying training times for condensation on CIFAR10 \textbf{(a)} and ImageNet20 \textbf{(b)}. The performance metrics for achieving convergence and their associated times for CIFAR10 \textbf{(c)} and ImageNet20 \textbf{(d)}. The diameters of the circles represent the time needed for convergence.}
    \label{time}
    \vspace{-8mm}
\end{figure}

\vspace{-0.5mm}
The efficiency of dataset condensation has a significant impact on training time and computational expenses. In order to assess the condensation efficiency, we measure the training time for condensing CIFAR-10 and the ImageNet-subset, utilizing an RTX-2080 Ti and RTX 3090, respectively. As depicted in Fig.~\ref{time}, IEM consistently outperforms IDC across different training times and IEM converges more quickly. Besides, although IDC slightly surpasses IEM on ImageNet20, this advantage is accompanied by substantially higher time and computational expenses (e.g., taking 500 hours for condensing ImageNet10 on RTX 3090). The high efficiency of IEM allows for its efficient deployment while minimizing increases in various operational burdens.

\vspace{-1mm}
\subsection{Generalization Comparison.} 
\begin{figure}[t] 
    \centering
    \includegraphics[width=0.85\columnwidth]{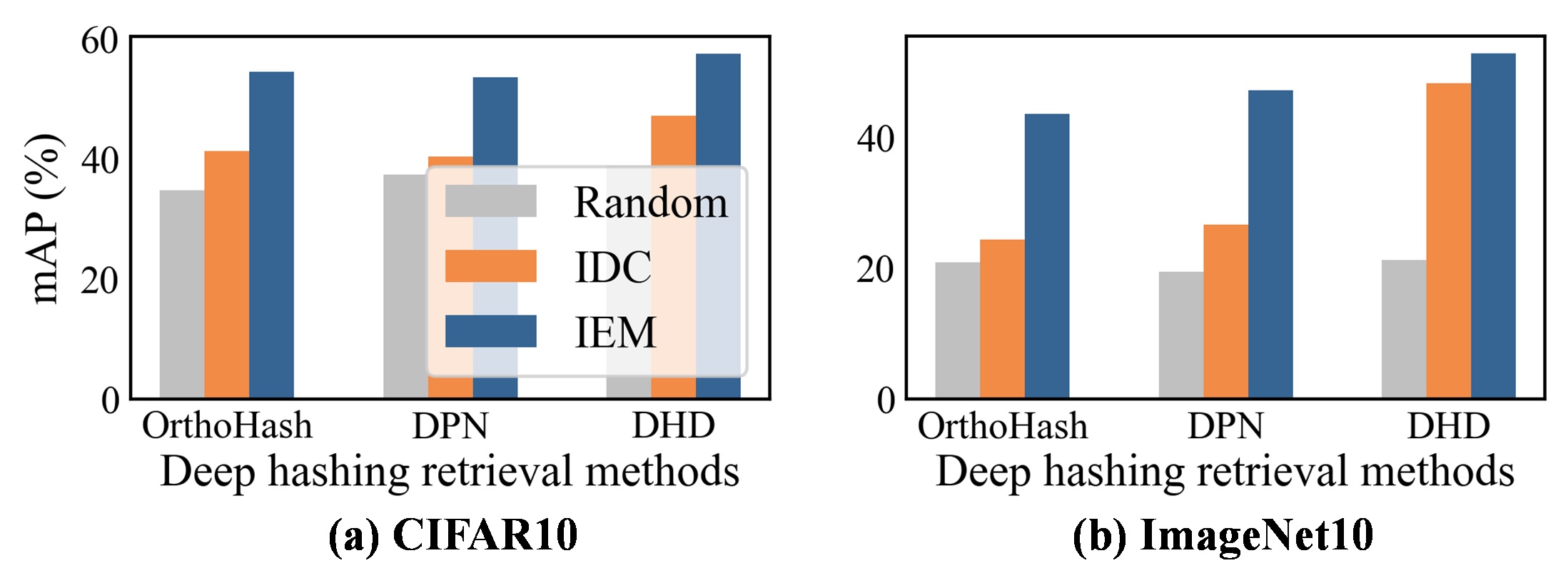}
    \vspace{-1.5mm}
    \captionsetup{font={footnotesize}}
    \caption{The generalization performance of IEM and IDC across DHR methods on CIFAR10 \textbf{(a)} and ImageNet10 \textbf{(b)}.}
    \label{different_ways}
    \vspace{-2mm}
\end{figure}

\vspace{-0.5mm}
To evaluate the extensive applicability of IEM, we compare its condensation performance using several methods of DHR, including DPN~\cite{fan2020deep}, OrthoHash~\cite{hoe2021one}, and DHD~\cite{jang2021deep}. As illustrated in Fig.~\ref{different_ways}, IEM demonstrates superior performance across all three DHR models in both CIFAR10 and ImageNet10. These results highlight IEM is compatible with current DHR methods, thereby expanding its potential for deployment in real-world retrieval systems.

\vspace{-1mm}
\subsection{Ablation Study.} 
\vspace{-0.5mm}
To evaluate the impact of two augmentation strategies in IEM, we perform ablation experiments on Network Augmentation (NA) and Dataset Augmentation (DA), respectively. The results shown in Table~\ref{abalation_1} indicate that the performance of IEM is significantly diminished without NA and DA. The concentration effect improves to varying degrees when either DA or NA is used in isolation, but the best results are obtained when both are employed simultaneously.

\begin{table}[t]
\centering
\captionsetup{font={footnotesize}}
\caption{The results of ablation study. The imgs/cls denotes the number of synthetic images per class.}
\vspace{-1.5mm}
\label{abalation_1}
\begin{tabular}{cccc}
\toprule
Dataset                                                                           & NA & DA & mAP            \\ \hline
\multirow{4}{*}{\begin{tabular}[c]{@{}c@{}}CIFAR10\\ (50 imgs/cls)\end{tabular}}  & \ding{55}  & \ding{55}  & 43.22          \\
                                                                                  & \checkmark  & \ding{55}  & 47.99          \\
                                                                                  & \ding{55}  & \checkmark  & 57.73          \\
                                                                                  &\checkmark  & \checkmark  & \textbf{57.88} \\ \midrule
\multirow{4}{*}{\begin{tabular}[c]{@{}c@{}}ImageNet10\\ (3imgs/cls)\end{tabular}} & \ding{55} & \ding{55}  & 25.83          \\
                                                                                  & \checkmark  & \ding{55}  & 30.23          \\
                                                                                  & \ding{55}  & \checkmark  & 47.63          \\
                                                                                  & \checkmark  & \checkmark & \textbf{50.38} \\ \bottomrule
\end{tabular}
\vspace{-6mm}
\end{table}

\section{Conclusion}
\vspace{-0.5mm}
In conclusion, we propose IEM, a novel approach that represents the first endeavor in condensing datasets for DHR. Extensive experimental results demonstrate significant improvements in both performance and efficiency. Notably, IEM integrates seamlessly with existing image hashing retrieval frameworks, enhancing the efficiency of real-world hashing retrieval systems.

\bibliography{references} 

\end{document}